\documentclass[10pt]{article} 
\usepackage[preprint]{tmlr}

\usepackage{hyperref}
\usepackage{url}

\usepackage[utf8]{inputenc}
\usepackage[T1]{fontenc}
\usepackage{hyperref}
\usepackage{url}
\usepackage{booktabs}
\usepackage{amsfonts}
\usepackage{amsmath}
\usepackage{amssymb}
\usepackage{nicefrac}
\usepackage[expansion=false]{microtype}
\usepackage{graphicx}
\usepackage{xcolor}
\usepackage{subcaption}
\usepackage{float}
\usepackage{multirow}
\usepackage{xspace}
\usepackage{longtable}
\usepackage[section]{placeins}

\newcommand{\ours}{\textsc{RocketPFN}\xspace}
\newcommand{\tabpfn}{\textsc{TabPFN}\xspace}
\newcommand{\rocket}{\textsc{Rocket}\xspace}
\newcommand{\hctwo}{\textsc{HC2}\xspace}

\title{RocketPFN: Accurate Time Series Classification via In-Context Learning}

\author{\name Franco Martino O'Rourke \quad\quad Ana Trisovic \quad\quad Dimitris Bertsimas}

\begin{document}

\maketitle

\begin{abstract}
We introduce \ours, a training-free pipeline for time series
classification that combines random convolutional feature extraction (\rocket) with in-context classification via a pretrained tabular foundation model (TabPFN v2.5).
On 92 UCR datasets (30-resample protocol), \ours matches \hctwo, the strongest published method on the archive, in mean accuracy (both 0.900, Wilcoxon $p=0.50$), with no training on the target data and a median inference time of 30 seconds per fold. It also significantly outperforms every individual classifier in the HC2 ensemble. On UEA (20 datasets) the difference is likewise not statistically significant.
A separate comparison concerns TSC foundation models: when paired with the same downstream classifier, MOMENT, Mantis, and MantisV2 are all significantly outperformed by \ours using fewer extracted features and no learned parameters ($p<0.001$ in each case). This holds even when the encoders were pretrained on corpora that include the UCR training samples. We propose this two-stage pipeline as a reference point for evaluating zero-shot TSC foundation models.

\end{abstract}

\section{Introduction}
\label{sec:intro}

Time series classification is a core problem across science and
engineering. Detecting seizures from EEG~\citep{chaovalitwongse2007brain},
recognizing human activity from wearable sensors~\citep{lara2013har},
monitoring industrial equipment for fault diagnosis~\citep{lei2020fault},
and classifying seismic events~\citep{arul2021applications} all reduce to
the same fundamental task: assigning a discrete label to a sequence of
measurements. The 
UCR~\citep{dau2019ucr} and UEA~\citep{bagnall2018uea} archives have 
provided a standardized benchmark of over 150 datasets spanning 
univariate and multivariate settings, and have driven steady, 
measurable progress on the problem over two decades.

The strongest published method on these archives is HIVE-COTE~2.0
(\hctwo)~\citep{middlehurst2021hivecote2}, a meta-ensemble of four
specialized classifiers, each itself an ensemble, covering shapelets,
dictionaries, interval-based features, and convolutions, combined via a
weighted probabilistic controller. \hctwo achieves the highest mean
accuracy on UCR and UEA, but at a steep computational cost, with
training times of hours to days per dataset ~\citep{middlehurst2021hivecote2}.

A separate family of methods reaches near-\hctwo accuracy at a small
fraction of its cost. \rocket~\citep{dempster2020rocket} and its
successors, map each time series to a fixed-length vector of convolutional summaries computed from randomly sampled kernels. These vectors are
deliberately overcomplete: default \rocket uses $10{,}000$ kernels to
produce $20{,}000$ features, often exceeding the number of training
examples. This makes a strongly regularised linear classifier the natural choice for extracting signal from the random feature space. Despite their simplicity, the accuracy gap between this family and \hctwo on UCR is small~\citep{middlehurst2024bakeoff}.

On a separate line, TSC foundation
models~\citep{goswami2024moment, feofanov2025mantis,
feofanov2026mantisv2} invest substantial design effort, pretraining
data, and compute in the feature extraction step. The motivation is to
learn embeddings of moderate dimensionality (typically 256 to 1{,}024)
that improve quality and compactness,
enabling more expressive non-linear downstream classifiers and,
ultimately, higher accuracy. Used as frozen feature
extractors, these models do not currently match \hctwo or the
random-kernel family on UCR or UEA.

Meanwhile, a different class of classifiers has matured in the tabular 
machine learning community. \tabpfn~\citep{hollmann2023tabpfn,
hollmann2025tabpfn} is a Prior-data Fitted Network~\citep{muller2022pfns},
pretrained on millions of synthetic datasets to approximate Bayesian
inference via in-context learning: it receives a labeled training set and
unlabeled test points as input and predicts class probabilities in a
single forward pass, with no gradient updates on the target data. The version we build on,
\tabpfn~v2.5~\citep{priorlabs2025tabpfn25}, scales to 50{,}000 samples
and 2{,}000 features, outperforming tuned gradient-boosted trees on 
standard tabular benchmarks. Crucially, like Rocket, TabPFN requires no fitting on the target task, making it well suited to small-data regimes, but unlike the linear classifiers typically used after Rocket, it can represent non-linear decision rules.

This raises a natural question. Can a training-free pipeline, combining
random convolutional features with in-context tabular classification,
reach the accuracy of the methods that dominate these archives?

We answer this question experimentally. We introduce \ours, a 
two-stage, training-free pipeline: random convolutional feature 
extraction (\`{a} la \rocket) followed by in-context classification via 
\tabpfn~v2.5. The design is constrained by a practical mismatch---\rocket 
produces 20{,}000 features from 10{,}000 kernels, while \tabpfn is 
designed for up to 2{,}000---which we resolve by splitting kernels into 
$G$ groups of $1{,}000$ and ensembling the resulting probability 
estimates. No parameters are fitted to the target data at any stage.

On 92 datasets of the UCR archive (30-resample protocol), \ours ($G{=}10$) 
achieves the same mean accuracy as \hctwo (both 0.900, Wilcoxon 
$p = 0.50$)~\citep{wilcoxon1945}, while requiring only a median of 30 
seconds per fold on a GPU. It significantly outperforms every individual 
classifier in the \hctwo ensemble, as well as \rocket, InceptionTime, 
HC1, and TS-CHIEF~\citep{fawaz2020inceptiontime, lines2018hivecote, 
shifaz2020tschief}. Even at one-tenth the kernel budget ($G{=}1$, 
$1{,}000$ kernels), \ours already significantly outperforms all four 
\hctwo components and \rocket on UCR.

A second comparison concerns the zero-shot use of TSC foundation models. When 
MOMENT~\citep{goswami2024moment}, Mantis~\citep{feofanov2025mantis}, and 
MantisV2~\citep{feofanov2026mantisv2} are used as zero-shot feature 
extractors and paired with the same \tabpfn downstream classifier, all 
three are significantly outperformed by \ours at $G{=}1$ ($p < 0.001$ in 
each case, 103 UCR datasets). This holds even when the encoders were pretrained on corpora that include the UCR training samples. It also holds
when \ours is restricted to $k{=}100$ kernels and 200 features, fewer
than the embedding dimension of any of the three encoders, ruling out
dimensionality as a confound.

We propose this two-stage pipeline---random convolutional features
classified in context by a pretrained tabular foundation
model---as a reference point for zero-shot TSC foundation models.
The pretraining cost and design complexity required to learn
low-dimensional embeddings should be justified by concrete conditions
under which those embeddings outperform this training-free baseline.

Our main contributions are as follows:

\begin{enumerate}
    \item \textbf{A training-free pipeline that matches the state of the 
    art.} \ours ($G{=}10$) achieves the same mean accuracy as \hctwo on 
    UCR (both 0.900, $p = 0.50$) and is not significantly different on 
    UEA, with no parameter fitting on the target data and a median 
    wall-clock time of 30 seconds per fold 
    (Sections~\ref{sec:vs_sota}--\ref{sec:efficiency}).

    \item \textbf{In-context learning outperforms ridge regression on 
    random convolutional features.} At the same $10{,}000$-kernel budget, replacing 
    ridge regression with \tabpfn yields a significant accuracy gain on 
    UCR, confirming that the classifier is a non-trivial contributor 
    (Section~\ref{sec:vs_sota}).

    \item \textbf{Temporal encoding is necessary.} \tabpfn applied to 
    raw flattened time series fails completely on trivially solvable tasks
    (Section~\ref{sec:feature_extraction}).

    \item \textbf{Random convolutional features as a reference point for
    zero-shot TSC foundation models.} \ours ($G{=}1$) significantly 
    outperforms MOMENT, Mantis, and MantisV2---each paired with the same 
    downstream classifier---on 103 UCR datasets, including models 
    pretrained on the UCR training splits themselves 
    (Section~\ref{sec:vs_pretrained}).
\end{enumerate}

%
\section{Related Work}
\label{sec:background}

A time series classification dataset consists of $n$ instances $\{(\mathbf{X}_i, y_i)\}_{i=1}^{n}$, where each $\mathbf{X}_i \in \mathbb{R}^{m \times t}$ is a (possibly multivariate) series with $m$~channels and $t$~time steps, and $y_i \in \{1, \ldots, C\}$ is a class label.
When $m = 1$ the problem is univariate; the UEA archive focuses on the multivariate case, where $m > 1$.

\paragraph{Random convolutional features.}
\rocket~\citep{dempster2020rocket} transforms each time series into a fixed-length feature vector by applying a large number of random convolutional kernels.
Each kernel has a randomly sampled length $l \in \{7, 9, 11\}$, weights drawn from $\mathcal{N}(0, 1)$ and mean-centered, bias from $\mathcal{U}(-1, 1)$, and dilation $d = \lfloor 2^x \rfloor$ with $x \sim \mathcal{U}(0, A)$, where $A = \log_2 \frac{t-1}{l-1}$, so that the effective receptive field can span the full series length.
Padding is applied with probability $1/2$; stride is always one.
Each kernel is convolved with the input to produce a feature map, from which two summary statistics are extracted: the global maximum and the proportion of positive values (PPV).
With $k$~kernels this yields a $2k$-dimensional feature vector per instance.
The default configuration uses $k = 10{,}000$ kernels and feeds the resulting 20{,}000 features to a ridge regression classifier.

For multivariate series ($m > 1$), each kernel is assigned a randomly selected subset of $K$~channels, where $K = \lfloor 2^u \rfloor$ with $u \sim \mathcal{U}(0, \log_2(\min(m, l) + 1))$, so most kernels focus on a small number of channels.
Each selected channel receives its own weight vector of length $l$, drawn independently from $\mathcal{N}(0, 1)$ and mean-centered---yielding a kernel that acts as a proper multi-channel convolution restricted to the selected channels.
At each position, the outputs across all selected channels are summed into a single scalar; max and PPV are then computed over this summed signal.
Each kernel therefore still produces exactly two features regardless of the number of channels or the size of the selected subset.

The key insight behind \rocket is that while any single random kernel captures only a weak signal, the aggregate of thousands of diverse kernels---spanning different lengths, dilations, and weight patterns---produces a rich, discriminative representation.
Dilation is particularly important: it allows short kernels to detect patterns at multiple temporal scales, from fine-grained local motifs to broad trends.

\paragraph{In-context learning for tabular data.}
\tabpfn~\citep{hollmann2023tabpfn,hollmann2025tabpfn} is a Prior-data Fitted Network~\citep{muller2022pfns}---a transformer pre-trained on synthetic tabular datasets sampled from structural causal models.
At inference time it receives the entire labeled training set and unlabeled test points as context, and outputs predicted class probabilities in a single forward pass approximating the Bayesian posterior predictive distribution $p(y \mid \mathbf{x}, \mathcal{D}_\text{train})$ without any gradient updates on the target data.
\tabpfn~v2.5~\citep{priorlabs2025tabpfn25} extends this to 50{,}000 training samples and is optimized for up to 2{,}000 features, using a deeper architecture (24~layers for classification), larger feature group size, and 64~learned thinking rows that provide additional computational capacity.
Predictions are aggregated across an internal ensemble of $e = 8$~estimators, each applying a distinct combination of preprocessing transformations (robust scaling, soft clipping, quantile normalization, standard scaling), random column and label-index shuffling, and a random subsample of 500~features; some estimators additionally augment the feature set with SVD components.
Concurrent with this work, \tabpfn~v3~\citep{grinsztajn2026tabpfn3technicalreport} was released, raising the supported number of samples, features, and classes (which extends coverage of the UCR and UEA archives), alongside changes to the architecture and training prior.
We use \tabpfn~v2.5 throughout; a brief assessment of \ours under \tabpfn~v3 is reported in Appendix~\ref{app:tabpfn3}.

\paragraph{Time series classification foundation models.}
An alternative to hand-crafted features is to learn temporal representations via large-scale pretraining.
MOMENT~\citep{goswami2024moment}, Mantis~\citep{feofanov2025mantis}, and MantisV2~\citep{feofanov2026mantisv2} are time series foundation models that can be used as zero-shot feature extractors for classification: each is pretrained on real or synthetic time series data and used as a zero-shot feature extractor, mapping each series through a frozen encoder to a fixed-length embedding (256--1\,024 dimensions) that is then passed to a downstream classifier.
MOMENT is pretrained on a large public corpus that includes the UCR and UEA training splits, among others, using a masked reconstruction objective.
Mantis is pretrained on a union of public time series collections that includes UCR and UEA training splits, using a contrastive learning objective.
Neither model is exposed to test split or labels at any stage.
MantisV2 uses synthetic data generated via CauKer~\citep{xie2026cauker}.

\paragraph{State-of-the-art ensembles.}
\hctwo~\citep{middlehurst2021hivecote2} is a meta-ensemble of four classifiers, each targeting a different representation domain: the Shapelet Transform Classifier (STC), the Temporal Dictionary Ensemble (TDE), the Diverse Representation Canonical Interval Forest (DrCIF), and Arsenal---itself an ensemble of \rocket classifiers.
A weighted probabilistic controller (CAWPE) combines their predictions.
\hctwo is currently the most accurate method on the UCR and UEA archives, significantly outperforming all individual classifiers including \rocket, InceptionTime, and TS-CHIEF.

\section{Method}
\label{sec:method}

\ours is a two-stage pipeline: random convolutional feature extraction followed by in-context classification.

Given a dataset $\{(\mathbf{X}_i, y_i)\}_{i=1}^{n}$ with $\mathbf{X}_i \in \mathbb{R}^{m \times t}$, each instance is first $z$-normalised along the time axis to zero mean and unit variance---the standard preprocessing recommended for \rocket~\citep{dempster2020rocket} and applied by default in the aeon toolkit~\citep{middlehurst2024aeon}, which we use for all \rocket transforms and dataset loading.

We then generate $G$ independent groups of $k = 1{,}000$ random convolutional kernels (parameterized as in standard \rocket~\citep{dempster2020rocket}).
Each group $g$ transforms every instance into a $2k = 2{,}000$-dimensional feature vector, producing a feature matrix $\mathbf{F}_g \in \mathbb{R}^{n \times 2000}$ that aligns with \tabpfn~v2.5's recommended feature range.

Each feature matrix is passed to a \tabpfn~v2.5 classifier, which returns predicted probabilities $\mathbf{P}_g \in \mathbb{R}^{n_\text{test} \times C}$ via in-context learning (no gradient updates on the target data).
The final prediction averages across groups:
\begin{equation}
    \hat{y}_j = \arg\max_{c}\ \frac{1}{G} \sum_{g=1}^{G} P_g[j, c].
    \label{eq:ensemble}
\end{equation}

This design is motivated by a simple constraint: \tabpfn~v2.5 is designed for up to $2{,}000$ features, while standard \rocket uses $10{,}000$ kernels (producing $20{,}000$ features).
Rather than truncating or compressing features, we distribute the kernels across $G$~groups and ensemble the classifiers---a strategy that naturally provides diversity, since each group operates on a different random projection of the time series.

With $G = 10$ groups, \ours uses the same total number of random kernels as default \rocket ($10 \times 1{,}000 = 10{,}000$), enabling a controlled comparison that isolates the classifier: ridge regression versus \tabpfn.
The entire pipeline is \emph{training-free}: the convolutional kernels are randomly generated (not learned), and \tabpfn classifies via its pre-trained weights.
No parameters are fitted to the target dataset.

\section{Experiments}
\label{sec:experiments}

\subsection{Datasets and Evaluation Protocol}
\label{sec:protocol}

We evaluate \ours ($G{=}1$ and $G{=}10$) on the UCR archive~\citep{dau2019ucr} (univariate) and the UEA archive~\citep{bagnall2018uea} (multivariate), restricting to datasets within \tabpfn's operational scope ($C \leq 10$ classes, $n_\text{train} \leq 50{,}000$), yielding 103~UCR and 23~UEA datasets (126 total).
Of the 128~UCR and 30~UEA datasets in the archives, 25~UCR and 7~UEA are excluded, all due to the class limit ($C$ ranging from 11 to 60); no dataset in the archive exceeds the training-size limit.
Excluded datasets are listed in Appendix~\ref{app:excluded}.
Each pairwise comparison is restricted to the intersection of datasets for which both methods have results: baseline accuracies for \hctwo and its components are taken from~\citet{middlehurst2021hivecote2}, who evaluated on a standard benchmark of 112~UCR and 26~UEA datasets (variable-length and other incompatible datasets excluded), so comparisons against those methods use that intersection.

All methods follow the same evaluation protocol: 30~stratified resamples per dataset, where resample~0 is the original archive train/test split and resamples~1--29 are generated by stratified shuffling with seeds~1--29, preserving class proportions.
We report mean accuracy over the 30 resamples.
Pairwise comparisons use the Wilcoxon signed-rank test~\citep{wilcoxon1945} ($p < 0.05$); rankings across multiple methods are summarized with critical difference (CD) diagrams~\citep{demsar2006statistical}.
Encoder results (MOMENT, Mantis, MantisV2) are computed by us under this protocol; all other baseline results are taken from~\citet{middlehurst2021hivecote2}.
All \ours runs use the \texttt{aeon} toolkit~\citep{middlehurst2024aeon} for dataset loading and feature extraction.
Per-dataset results are in Appendix~\ref{app:full_results}.

\subsection{Temporal Encoding is Necessary}
\label{sec:feature_extraction}

We first ask whether \tabpfn can classify time series \emph{without} any temporal encoding, by treating each univariate series of length $t$ as a flat vector of $t$ raw values and passing it directly to \tabpfn.
Table~\ref{tab:flat_vs_sota} reports the comparison on the 90 UCR datasets that satisfy both $t \leq 2{,}000$, the feature range for which TabPFN is
designed, and the HC2 bake-off benchmark for which published baseline numbers are available.
The answer is clear: despite being state-of-the-art on tabular benchmarks, \tabpfn~flat is significantly outperformed by \emph{every} established classifier on the UCR archive---including \rocket---and is statistically tied only with the two weakest baselines (Table~\ref{tab:flat_vs_sota}).

\begin{table}[htbp]
\centering
\caption{\textbf{\tabpfn (flattened) vs.\ published baselines} on 90 UCR datasets (Wilcoxon signed-rank test, 30-resample protocol).
\tabpfn~flat is significantly worse than all methods except TDE and STC.}
\label{tab:flat_vs_sota}
\small
\begin{tabular}{lccc}
\toprule
\textbf{Method} & \textbf{Mean Acc.} & \textbf{vs.\ TabPFN flat} & \textbf{$p$-value} \\
\midrule
HC2           & 0.900 & $+0.053$ & $<$0.001 *** \\
TS-CHIEF      & 0.884 & $+0.037$ & $<$0.001 *** \\
HC1           & 0.891 & $+0.044$ & $<$0.001 *** \\
InceptionTime & 0.882 & $+0.035$ & 0.005 ** \\
ROCKET        & 0.882 & $+0.034$ & $<$0.001 *** \\
Arsenal       & 0.882 & $+0.035$ & $<$0.001 *** \\
DrCIF         & 0.883 & $+0.036$ & $<$0.001 *** \\
TDE           & 0.869 & $+0.022$ & 0.138 n.s. \\
STC           & 0.871 & $+0.023$ & 0.446 n.s. \\
\midrule
\textbf{TabPFN flat} & 0.847 & --- & --- \\
\bottomrule
\end{tabular}
\end{table}

The failure is not merely quantitative---it is structural.
\tabpfn is a powerful classifier, but without any inductive bias toward temporal structure it has no way to recover this information.
\rocket features restore it by construction: convolutional max-pooling captures \emph{whether} a pattern occurs regardless of \emph{where} in the series, and PPV captures its prevalence---both of which are invariant to translation and robust to noise.

\begin{figure}[htbp]
    \centering
    \includegraphics[width=0.95\linewidth]{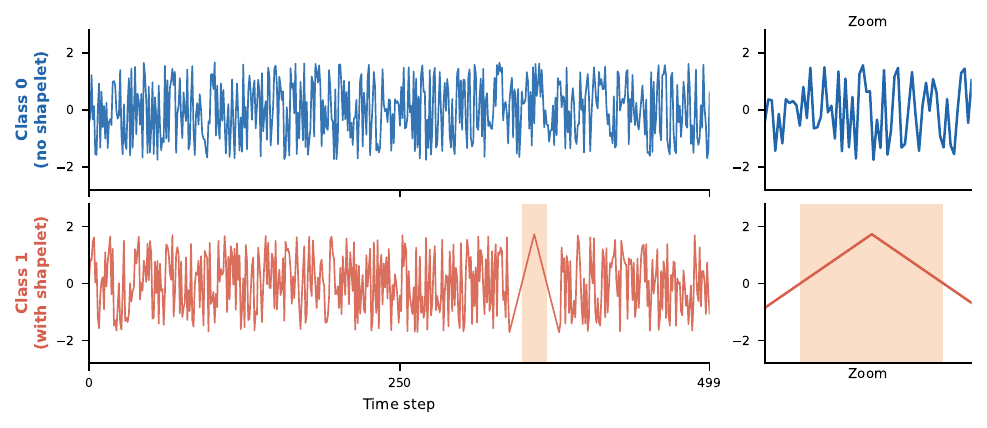}
    \caption{\textbf{ShapeletSim: trivial for humans, impossible for flattened features.}
    Class~0 is pure Gaussian noise; Class~1 contains a triangular pulse at a random position (highlighted in orange, enlarged right).
    \tabpfn~flat: $\approx 0.50$ (chance).
    \ours ($G{=}1$): $\approx 1.00$ via translation-invariant pooling.}
    \label{fig:shapelet_sim}
\end{figure}

Figure~\ref{fig:shapelet_sim} illustrates the failure mode with a concrete example.
ShapeletSim ($t = 500$, $n_\text{train} = 20$) has two classes: both are Gaussian noise, but Class~1 contains a short triangular pulse at a \emph{random} position.
The pattern is immediately visible to a human, yet \tabpfn~flat scores $\approx 0.50$---random chance.
Encoding the same series with a single group of 1{,}000 \rocket kernels recovers near-perfect accuracy ($\approx 1.00$).

\subsection{In-Context Learning Outperforms Ridge Regression on Random Convolutional Features}
\label{sec:vs_rocket}

Before turning to the full state-of-the-art comparison, we isolate the contribution of the downstream classifier.
At an \emph{identical} kernel budget of $10{,}000$ random kernels, \ours ($G{=}10$, $k{=}1{,}000$) and \rocket differ only in the classifier applied to the resulting features: \tabpfn in-context learning versus ridge regression with built-in cross-validated regularisation.
On UCR, replacing ridge with \tabpfn yields a significant accuracy gain of $+0.018$ (0.900 vs.\ 0.882, $p < 0.001$; Table~\ref{tab:vs_sota}, \rocket row), with \ours winning 70 datasets, tying on 3 and losing on 19.
The significant gain confirms that in-context learning extracts more discriminative information from the same random features than a regularised linear model can: \tabpfn inherits \rocket's temporal inductive bias while replacing ridge's linear decision boundary with non-linear, context-dependent posterior predictions.

The effect is strong enough that \ours remains competitive at a fraction of the kernel budget.
With $G{=}1$ (only $1{,}000$ kernels, a tenth of \rocket's default), \ours already surpasses \rocket in mean accuracy on UCR (0.894 vs.\ 0.882) and is not significantly different from it on UEA ($p = 0.70$; same $10\times$ feature reduction).

\subsection{\ours Matches the State of the Art}
\label{sec:vs_sota}

\begin{figure}[htbp]
\centering
\begin{subfigure}[t]{0.99\linewidth}
    \centering
    \includegraphics[width=\linewidth]{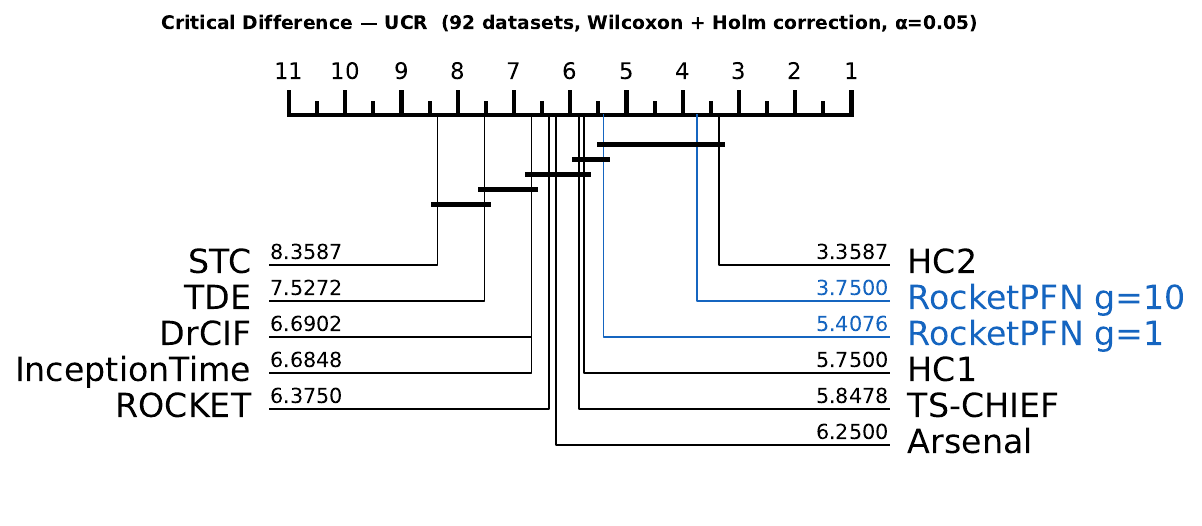}
    \caption{UCR (92 datasets)}
\end{subfigure}
\begin{subfigure}[t]{0.99\linewidth}
    \centering
    \includegraphics[width=\linewidth]{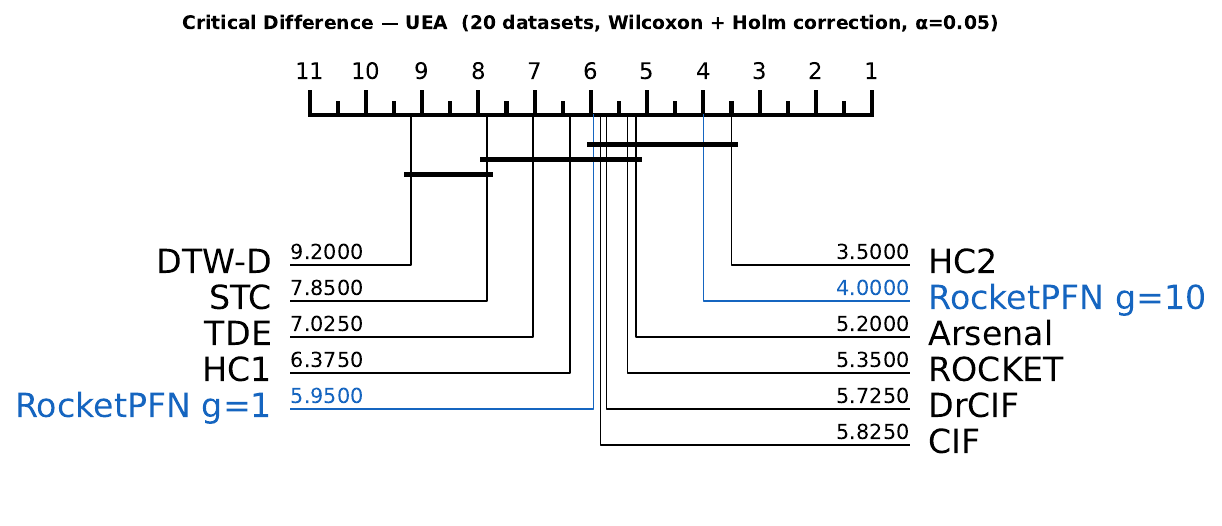}
    \caption{UEA (20 datasets)}
\end{subfigure}
\caption{\textbf{Critical difference diagrams} (30-resample protocol, Wilcoxon + Holm correction, $\alpha = 0.05$).
Thick bars connect methods with no significant difference.
On UCR (92 datasets), \ours ($G{=}10$) and \hctwo are in the same clique; all other methods are ranked strictly below.
On UEA (20 datasets), the small benchmark limits statistical power; \ours ($G{=}10$) forms a single clique with \hctwo and most other methods, consistent with competitive performance across multivariate datasets.}
\label{fig:cd_diagrams}
\end{figure}

On UCR ($n = 92$), \ours ($G{=}10$) achieves the same mean accuracy as \hctwo (both 0.900, $p = 0.50$): a two-component, training-free pipeline reaches the same mean accuracy as a meta-ensemble of four specialised classifiers, each requiring hours of training.
On UEA ($n = 20$), the benchmark is too small to draw strong statistical conclusions---with only 20 datasets, even sizable differences often fail to reach significance---but \ours ($G{=}10$) achieves comparable mean accuracy to \hctwo (both 0.747 at three decimal places, with \hctwo ahead by less than 0.001, $p = 0.37$) and ranks near the top of all methods evaluated.
Figure~\ref{fig:cd_diagrams} shows the complete ranking; on UCR, \ours and \hctwo form a single clique and every other method falls strictly below.

\ours ($G{=}10$) does not merely match \hctwo---it significantly outperforms every individual classifier in the \hctwo ensemble: Arsenal, DrCIF, TDE, and STC, as well as InceptionTime, HC1, and TS-CHIEF ($p < 0.005$ on UCR for all; Table~\ref{tab:vs_sota}).

\begin{table}[htbp]
\centering
\caption{\textbf{Pairwise comparisons on UCR} (Wilcoxon signed-rank test, 30-resample protocol) between \ours and published baselines.
Mean accuracy over 92 UCR datasets. W/T/L from the perspective of \ours ($G{=}10$).}
\label{tab:vs_sota}
\small
\begin{tabular}{lccccl}
\toprule
\textbf{Method} & $n$ & \textbf{Mean Acc.} & \textbf{W/T/L vs.\ \ours ($G{=}10$)} & \textbf{$p$-value} & \textbf{Sig.} \\
\midrule
\ours ($G{=}10$)  & 92 & 0.900 & --- & --- & --- \\
\ours ($G{=}1$)   & 92 & 0.894 & --- & --- & --- \\
\midrule
HC2           & 92 & 0.900 & 44/3/45  & 0.504 & n.s. \\
HC1           & 92 & 0.891 & 57/2/33  & 0.004 & ** \\
TS-CHIEF      & 92 & 0.884 & 57/2/33  & 0.003 & ** \\
DrCIF         & 92 & 0.884 & 66/1/25  & $<$0.001 & *** \\
InceptionTime & 92 & 0.880 & 69/2/21  & $<$0.001 & *** \\
Arsenal       & 92 & 0.882 & 69/3/20  & $<$0.001 & *** \\
ROCKET        & 92 & 0.882 & 70/3/19  & $<$0.001 & *** \\
STC           & 92 & 0.871 & 75/2/15  & $<$0.001 & *** \\
TDE           & 92 & 0.870 & 65/2/25  & $<$0.001 & *** \\
\bottomrule
\end{tabular}
\end{table}

The result holds even at a tenth of the kernel budget.
With $G{=}1$ ($1{,}000$ kernels, 2{,}000 features), \ours already significantly outperforms all four \hctwo components and \rocket on UCR ($p < 0.001$), and is not significantly different from \hctwo.
On UEA, the 20-dataset benchmark limits statistical power throughout, and $G{=}10$ outperforms \rocket ($p = 0.033$); both configurations rank near the top across UEA datasets.
On UCR, a single group of $1{,}000$ random kernels already captures the temporal structure needed for competitive performance; $G{=}10$ closes the remaining gap with the full ensemble.

\FloatBarrier
\subsection{Random Features Outperform TSC Foundation Models}
\label{sec:vs_pretrained}

We now ask whether a TSC foundation model, used as a zero-shot feature extractor and paired with the same \tabpfn classifier, can surpass random convolutional features.
To rule out feature dimensionality as a confound from the start, we compare \ours restricted to $G{=}1$, $k{=}100$ kernels (only $200$ features---fewer than any of the three encoders, whose embedding dimension ranges from $256$ (Mantis) to $1{,}024$ (MOMENT)) against each encoder's full-dimensional representation.
Under this stringent budget, \ours still significantly outperforms all three encoders on 103 UCR datasets ($p < 0.001$ in each case; Table~\ref{tab:vs_pretrained}), winning 64--77 of 103 datasets.

\begin{table}[htbp]
\centering
\caption{\textbf{RocketPFN vs.\ TSC foundation models} (Wilcoxon signed-rank test, 30-resample protocol).
All TSC foundation models are used as zero-shot feature extractors with \tabpfn as the downstream classifier.
$n = 103$ UCR datasets. W/T/L and $p$-value are from the perspective of \ours ($G{=}1$, $k{=}100$), which uses only 200 features---fewer than any of the three encoders---to rule out dimensionality as a confound.}
\label{tab:vs_pretrained}
\small
\begin{tabular}{lccccl}
\toprule
\textbf{Encoder} & \textbf{Dim} & \textbf{Mean Acc.} & \textbf{W/T/L } & \textbf{$p$-value} & \textbf{Sig.} \\
\midrule
\ours ($G{=}1$, $k{=}1\,000$)  & 2\,000 & 0.887 & --- & --- & --- \\
\ours ($G{=}1$, $k{=}100$)     &   200  & 0.874 & --- & --- & --- \\
\ours ($G{=}10$, $k{=}1\,000$) & 2\,000 & 0.894 & --- & --- & --- \\
\midrule
Mantis + \tabpfn    & 256   & 0.859 & 64/2/37  & $<$0.001 & *** \\
MantisV2 + \tabpfn  & 512   & 0.851 & 69/2/32  & $<$0.001 & *** \\
MOMENT + \tabpfn    & 1\,024 & 0.845 & 77/1/25  & $<$0.001 & *** \\
\bottomrule
\end{tabular}
\end{table}

The finding is especially striking since Mantis and MOMENT were both pretrained on corpora that include the UCR training splits, eliminating pretraining-data mismatch as an explanation for the gap.

The gap is not an artifact of the downstream classifier, either: in the results reported by the respective papers, these encoders do not reach state-of-the-art accuracy on UCR under any of the classification heads they study (random forest, XGBoost, linear probe, fine-tuned MLP).

\FloatBarrier
\subsection{Computational Cost}
\label{sec:efficiency}

\ours requires \emph{zero training on the target data}: \rocket applies a fixed random transform and \tabpfn classifies via a single forward pass through a network pre-trained offline.
There is no fitting step, no gradient computation, and no hyperparameter search at deployment time.
On the 112~standard benchmark datasets (92 UCR + 20 UEA), \ours ($G{=}10$) completes in a median of \textbf{30.6~seconds per fold} (NVIDIA L40S GPU); with $G{=}1$ the median drops to \textbf{3.7~seconds per fold}.

\FloatBarrier
\subsection{Sensitivity Analysis}
\label{sec:sensitivity}

\ours has three core design choices: the random convolutional feature extractor, the number of groups $G$, and the feature width per group.
We evaluate three feature extractors: \rocket~\citep{dempster2020rocket}, which produces two features per kernel (global max and PPV); MiniRocket~\citep{dempster2021minirocket}, a faster near-deterministic variant designed to scale the kernel count to $50{,}000$; and MultiRocket~\citep{tan2022multirocket}, which targets higher accuracy with a richer pooling scheme.
Both MiniRocket and MultiRocket were designed with a much larger kernel budget in mind; when the feature width is fixed at $2{,}000$ per group (as required here by \tabpfn's input limit), the three extractors operate in a similar regime, and we do not expect large differences.
We vary the feature width per group in $\{672, 1344, 2016\}$ and $G \in \{1, 2, 5, 10, 25\}$, keeping $e{=}8$ fixed throughout, on the 92-dataset UCR benchmark.

\begin{figure}[htbp]
    \centering
    \includegraphics[width=0.64\linewidth]{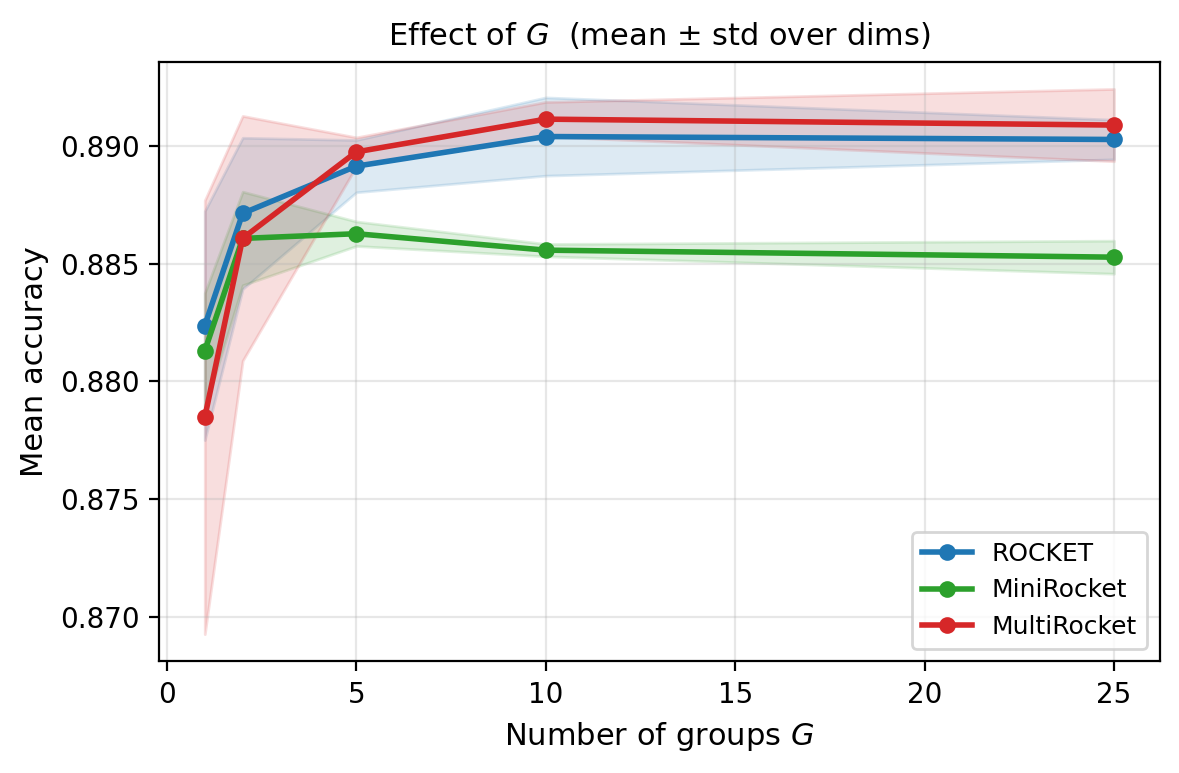}
    \caption{\textbf{Effect of $G$ and feature extractor} (92 UCR datasets).
    Lines show mean accuracy as a function of $G$ for \rocket, MiniRocket, and MultiRocket, averaged over feature widths $\{672, 1344, 2016\}$; shaded bands are $\pm$1 standard deviation.}
    \label{fig:ablation}
\end{figure}

Figure~\ref{fig:ablation} shows mean accuracy as a function of $G$ for each extractor, averaged over feature widths, with shaded bands indicating the standard deviation.
All three extractors behave similarly: accuracy rises steeply from $G{=}1$ to $G{\approx}5$ and plateaus thereafter, with differences between extractors below $0.006$ across the plateau.
The feature width has an equally muted effect across the range evaluated.
No significant differences emerge between extractors.
The same robustness extends to $e$: reducing \tabpfn's internal ensemble from $e{=}8$ to $e{=}1$ costs less than $0.003$ in mean accuracy once $G{\geq}5$, as diversity across groups largely subsumes within-group ensembling.

The chosen settings---\rocket, $G \in \{1, 10\}$, $k{=}1{,}000$ (width $2{,}000$)---were not the result of any optimisation or tuning.
$G{=}10$ matches \rocket's default kernel budget for a clean comparison; $G{=}1, k{=}100$ matches or stays below the embedding dimension of TSC foundation models.
Per-dataset selection of $G$ or feature width could yield further gains.

\FloatBarrier
%

\section{Conclusion}
\label{sec:conclusion}

\ours shows that a pipeline with no parameters fitted to the target
data can match \hctwo on UCR, by pairing
random convolutional features with in-context tabular classification.
The same pipeline outperforms three recent TSC foundation models used
zero-shot, even when restricted to fewer features than any of the
learned encoders produce.

We read these results as evidence that the bar for zero-shot temporal
representation learning is higher than current foundation models
suggest. Future foundation models should justify why their pretraining
cost and design complexity are worth it, at least by identifying concrete
conditions under which their learned embeddings outperform \ours, a
training-free pipeline of random convolutional features classified
in context.

Our findings complement concurrent work applying \tabpfn to time series forecasting~\citep{hoo2025tabpfnts}, and together suggest a broader research direction: tabular pre-fitted networks may serve as a general backbone for classification problems well beyond structured tabular data, provided suitable feature extractors bridge the gap between the source modality and the tabular interface---a recipe that could extend to audio, images, graphs, and multimodal settings alike.

\paragraph{Limitations}
\tabpfn~v2.5 supports up to $C{=}10$ classes, which excludes 25 UCR and 7 UEA datasets (class counts ranging from 11 to 60; full list in Appendix~\ref{app:excluded}).
This is a current, not a fundamental, limitation: any standard multi-class decomposition---one-vs-one, one-vs-rest, or error-correcting output codes~\citep{dietterich1994ecoc}---reduces a $C$-class problem to a collection of $\leq 10$-class subproblems compatible with \tabpfn, and future versions of \tabpfn may relax the bound directly.
We chose not to introduce a decomposition strategy here because it would mix our zero-shot result with an engineering choice that is not central to the paper's claim.
The intersection of the UEA bake-off benchmark in~\citet{middlehurst2021hivecote2} with TabPFN-compatible datasets contains only 20 datasets, which is near the practical lower bound for the Wilcoxon signed-rank test: even sizable per-dataset differences frequently fail to reach $\alpha = 0.05$.
We report the UEA comparison for completeness but treat it as inconclusive: \ours and \hctwo are tied in mean accuracy (both $0.747$) and statistically indistinguishable ($p = 0.37$).
\ours is evaluated on GPU while baseline methods are reported on CPU.
The reported timings should therefore not be read as a controlled hardware comparison, but rather as evidence of a structural gap: \ours requires no per-dataset training phase, so its wall-clock time is bounded by inference regardless of scale.


\bibliographystyle{plainnat}
\bibliography{refs}

\appendix
\section{Full Per-Dataset Results}
\label{app:full_results}

Tables~\ref{tab:ucr_results} and~\ref{tab:uea_results} report per-dataset mean accuracy (\%) over 30 resamples. Blank entries indicate the method is not applicable to that dataset (e.g.\ Flat is blank when $t > 2{,}000$; \hctwo is blank for datasets outside the standard benchmark of~\citet{middlehurst2021hivecote2}).

{\scriptsize
\setlength{\tabcolsep}{2.5pt}
\begin{longtable}{l ccccccc}
\caption{\textbf{UCR per-dataset results} (mean accuracy \%, 30 resamples). Flat = \tabpfn on raw flattened series ($t \leq 2{,}000$ only); Mantis, MantisV2, MOMENT each use the pretrained encoder with \tabpfn as downstream classifier. \textbf{Bold} indicates the best result per dataset (ties included). HC2 results were obtained from the published benchmark of~\citet{middlehurst2021hivecote2}; all other results were produced by the authors.}
\label{tab:ucr_results}\\
\toprule
Dataset & $G{=}1$ & $G{=}10$ & Flat & Mantis & MantisV2 & MOMENT & HC2 \\
\midrule
\endfirsthead
\multicolumn{8}{c}{{\tablename\ \thetable{} -- continued}} \\
\toprule
Dataset & $G{=}1$ & $G{=}10$ & Flat & Mantis & MantisV2 & MOMENT & HC2 \\
\midrule
\endhead
\midrule \multicolumn{8}{r}{{Continued on next page}} \\
\endfoot
\bottomrule
\endlastfoot
ACSF1 & 86.0 & \textbf{87.4}& 77.6 & 70.2 & 72.1 & 70.0 & 83.3 \\
AllGestureWiimoteX & 75.8 & \textbf{77.2}& 59.8 & 66.8 & 60.4 & 66.3 &  \\
AllGestureWiimoteY & 79.1 & \textbf{80.3}& 64.6 & 66.6 & 58.4 & 72.6 &  \\
AllGestureWiimoteZ & 73.2 & \textbf{75.0}& 54.9 & 67.7 & 60.0 & 62.9 &  \\
ArrowHead & 87.0 & \textbf{88.8}& 81.9 & 78.3 & 84.4 & 83.8 & 88.6 \\
BME & 99.7 & 99.7 & 98.4 & 99.2 & 81.5 & 96.1 & \textbf{100.0} \\
Beef & 76.3 & 77.0 & \textbf{87.0}& 60.1 & 63.2 & 74.3 & 79.7 \\
BeetleFly & 88.5 & 89.2 & 75.2 & 87.8 & 88.0 & \textbf{94.7}& 90.3 \\
BirdChicken & 92.2 & 92.2 & 78.3 & 87.8 & 89.7 & 87.2 & \textbf{95.2} \\
CBF & 99.9 & \textbf{100.0}& 94.3 & 99.9 & 99.7 & 98.4 & 99.8 \\
Car & 91.4 & \textbf{92.2}& 80.7 & 88.2 & 88.6 & 81.8 & 90.8 \\
Chinatown & 96.9 & 97.1 & \textbf{97.6}& 87.2 & 94.3 & 97.3 & 97.4 \\
ChlorineConcentration & 78.0 & 79.5 & \textbf{96.6}& 80.3 & 80.9 & 80.9 & 76.9 \\
CinCECGTorso & 86.9 & 88.5 & 91.3 & 74.7 & 75.4 & 73.1 & \textbf{99.8} \\
Coffee & \textbf{100.0}& \textbf{100.0}& \textbf{100.0}& 99.5 & \textbf{100.0}& 99.2 & 99.9 \\
Computers & 87.8 & \textbf{88.8}& 70.7 & 77.6 & 80.1 & 70.8 & 85.9 \\
DiatomSizeReduction & 95.4 & 95.6 & 95.3 & 90.5 & 90.8 & \textbf{95.8}& 92.2 \\
DistalPhalanxOutlineAgeGroup & 79.2 & 80.2 & 78.8 & 78.6 & 79.4 & 80.2 & \textbf{82.1} \\
DistalPhalanxOutlineCorrect & 83.7 & 83.9 & 82.9 & 82.5 & \textbf{84.7}& 83.1 & 82.8 \\
DistalPhalanxTW & 70.6 & 71.7 & 70.5 & \textbf{72.1}& 71.1 & 70.5 & 70.2 \\
DodgerLoopDay & 58.7 & \textbf{61.5}& 61.3 & 54.2 & 49.1 & 47.2 &  \\
DodgerLoopGame & 83.4 & \textbf{84.1}& 75.0 & 73.1 & 60.0 & 76.9 &  \\
DodgerLoopWeekend & 98.3 & 98.4 & \textbf{98.6}& 96.9 & 96.0 & 96.6 &  \\
ECG200 & 90.4 & \textbf{90.8}& 86.0 & 83.2 & 86.2 & 88.5 & 88.9 \\
ECG5000 & 94.7 & 94.7 & 94.3 & 94.1 & 94.3 & 94.6 & \textbf{94.8} \\
ECGFiveDays & 99.1 & 99.2 & 92.2 & 92.8 & 93.7 & 98.0 & \textbf{99.7} \\
Earthquakes & 74.5 & 74.8 & \textbf{74.9}& 74.8 & 74.8 & 73.6 & 74.8 \\
ElectricDevices & 91.1 & \textbf{91.6}& 85.5 & 89.3 & 88.4 & 81.7 & 89.8 \\
EthanolLevel & 84.8 & 86.6 & \textbf{94.2}& 60.8 & 61.4 & 80.7 & 83.6 \\
FaceFour & 91.5 & 92.2 & 90.4 & 95.8 & 93.6 & 87.4 & \textbf{96.7} \\
Fish & 97.4 & 97.6 & 91.7 & 95.0 & 94.9 & 94.5 & \textbf{98.2} \\
FordA & 94.8 & 95.1 & 92.6 & 91.2 & 92.9 & 92.9 & \textbf{95.4} \\
FordB & 92.7 & \textbf{93.1}& 89.8 & 89.7 & 90.4 & 91.1 & 93.0 \\
FreezerRegularTrain & 99.9 & 99.9 & 99.8 & 99.4 & 98.9 & 99.2 & \textbf{100.0} \\
FreezerSmallTrain & 99.1 & 99.3 & 98.3 & 95.2 & 93.7 & 87.9 & \textbf{99.8} \\
GesturePebbleZ1 & 97.2 & \textbf{97.5}& 92.3 & 96.1 & 95.3 & 93.8 &  \\
GesturePebbleZ2 & 97.4 & \textbf{98.0}& 92.5 & 96.5 & 96.0 & 94.3 &  \\
GunPoint & 99.4 & 99.5 & 96.0 & 99.2 & 98.3 & 97.9 & \textbf{99.9} \\
GunPointAgeSpan & 99.2 & 99.2 & 98.4 & 99.5 & \textbf{99.6}& 97.1 & \textbf{99.6} \\
GunPointMaleVersusFemale & 99.9 & 99.9 & 99.9 & 99.8 & 99.7 & 99.5 & \textbf{100.0} \\
GunPointOldVersusYoung & 99.1 & 99.1 & 99.9 & 99.9 & \textbf{100.0}& 97.9 & \textbf{100.0} \\
Ham & 84.7 & 85.7 & 84.6 & 77.1 & 77.2 & 79.3 & \textbf{85.9} \\
HandOutlines & \textbf{95.4}& 95.2 &  & 93.3 & 93.7 & 91.7 & 93.5 \\
Haptics & 54.2 & \textbf{55.7}& 48.2 & 52.2 & 51.6 & 48.4 & 54.7 \\
Herring & 62.2 & \textbf{63.5}& 61.0 & 63.0 & 59.8 & 60.4 & 62.6 \\
HouseTwenty & 96.1 & 96.2 & 86.1 & 95.7 & 93.6 & 90.3 & \textbf{98.6} \\
InlineSkate & 55.7 & \textbf{58.3}& 37.3 & 42.3 & 44.2 & 39.4 & 54.6 \\
InsectEPGRegularTrain & 98.9 & 98.9 & \textbf{100.0}& \textbf{100.0}& \textbf{100.0}& 93.7 & \textbf{100.0} \\
InsectEPGSmallTrain & 91.8 & 92.2 & \textbf{100.0}& \textbf{100.0}& \textbf{100.0}& 82.9 & 99.9 \\
ItalyPowerDemand & 95.9 & 96.2 & \textbf{96.4}& 91.4 & 91.2 & 95.5 & 96.2 \\
LargeKitchenAppliances & 94.8 & \textbf{95.1}& 73.6 & 88.7 & 82.9 & 85.9 & 93.5 \\
Lightning2 & 81.4 & \textbf{83.6}& 72.3 & 81.5 & 75.7 & 80.1 & 78.4 \\
Lightning7 & 80.4 & \textbf{82.0}& 67.5 & 77.0 & 73.7 & 74.7 & 79.8 \\
Mallat & 97.3 & 97.4 & 96.8 & 96.6 & 95.2 & 96.1 & \textbf{97.9} \\
Meat & 98.8 & 98.8 & 99.7 & 99.1 & 99.6 & \textbf{99.8}& 99.1 \\
MedicalImages & 83.6 & \textbf{84.6}& 81.2 & 77.3 & 78.7 & 80.4 & 81.0 \\
MelbournePedestrian & 90.8 & 91.3 & \textbf{97.9}& 94.4 & 95.8 & 89.3 &  \\
MiddlePhalanxOutlineAgeGroup & 62.9 & 65.4 & 70.5 & 70.5 & \textbf{71.8}& 69.9 & \textbf{71.8} \\
MiddlePhalanxOutlineCorrect & 84.7 & \textbf{85.2}& 85.1 & 82.8 & 83.9 & 85.0 & 83.6 \\
MiddlePhalanxTW & 56.3 & 58.2 & 58.4 & 57.5 & 58.0 & 58.1 & \textbf{59.2} \\
MixedShapesRegularTrain & 97.0 & 97.2 & 94.3 & 96.5 & 95.7 & 92.8 & \textbf{97.3} \\
MixedShapesSmallTrain & 93.8 & 94.3 & 84.2 & 92.3 & 91.1 & 84.2 & \textbf{95.2} \\
MoteStrain & 90.3 & 91.1 & 86.7 & 88.2 & 90.1 & 87.9 & \textbf{93.3} \\
OSULeaf & 95.8 & 96.4 & 67.1 & 93.7 & 92.7 & 84.0 & \textbf{96.6} \\
OliveOil & 89.4 & 90.0 & 90.9 & 90.0 & \textbf{91.7}& 88.0 & 90.3 \\
PhalangesOutlinesCorrect & 86.4 & \textbf{86.6}& \textbf{86.6}& 84.0 & 86.1 & 86.0 & 84.5 \\
PickupGestureWiimoteZ & 76.3 & 78.7 & 69.1 & \textbf{80.4}& 78.7 & 70.5 &  \\
Plane & \textbf{100.0}& \textbf{100.0}& 98.9 & \textbf{100.0}& \textbf{100.0}& 99.6 & \textbf{100.0} \\
PowerCons & 95.8 & 96.2 & \textbf{100.0}& 93.1 & 92.6 & 90.7 & 96.9 \\
ProximalPhalanxOutlineAgeGroup & 83.5 & 84.3 & 85.1 & 84.3 & 84.0 & 84.6 & \textbf{85.4} \\
ProximalPhalanxOutlineCorrect & 90.6 & \textbf{91.0}& 90.6 & 89.1 & 90.1 & 89.8 & 89.6 \\
ProximalPhalanxTW & 80.9 & 81.5 & \textbf{82.3}& 79.9 & 80.9 & 80.7 & 81.1 \\
RefrigerationDevices & 80.6 & \textbf{81.3}& 54.5 & 69.5 & 71.6 & 66.9 & 76.7 \\
Rock & 74.8 & 76.5 &  & 77.6 & 77.3 & 66.8 & \textbf{88.9} \\
ScreenType & 71.1 & \textbf{72.6}& 48.9 & 58.4 & 58.0 & 53.3 & 70.7 \\
SemgHandGenderCh2 & 95.9 & \textbf{96.5}& 96.0 & 93.0 & 87.2 & 87.9 & 96.1 \\
SemgHandMovementCh2 & 86.0 & \textbf{87.9}& 82.7 & 80.2 & 72.3 & 62.2 & 86.2 \\
SemgHandSubjectCh2 & 95.0 & \textbf{95.8}& 95.3 & 91.1 & 83.0 & 85.4 & 93.9 \\
ShakeGestureWiimoteZ & 88.1 & 89.4 & 76.7 & \textbf{90.1}& 89.3 & 88.7 &  \\
ShapeletSim & 99.9 & \textbf{100.0}& 49.6 & 92.5 & 80.0 & 92.4 & \textbf{100.0} \\
SmallKitchenAppliances & 84.6 & \textbf{85.1}& 75.4 & 82.7 & 81.2 & 70.4 & 84.1 \\
SmoothSubspace & 96.5 & 97.5 & \textbf{100.0}& 96.4 & 97.4 & 81.4 & 98.3 \\
SonyAIBORobotSurface1 & 95.5 & \textbf{95.9}& 86.4 & 89.5 & 89.3 & 90.2 & 95.2 \\
SonyAIBORobotSurface2 & 92.5 & 93.3 & 86.8 & 84.7 & 88.7 & 89.0 & \textbf{94.7} \\
StarLightCurves & 98.2 & \textbf{98.3}& 97.7 & 98.0 & 98.1 & 96.9 & 98.2 \\
Strawberry & 98.2 & 98.3 & \textbf{98.5}& 97.4 & 97.9 & 98.0 & 97.9 \\
Symbols & 97.0 & \textbf{97.1}& 85.9 & 95.7 & 93.9 & 95.4 & \textbf{97.1} \\
SyntheticControl & 99.6 & \textbf{99.8}& 98.8 & 99.4 & 99.1 & 99.1 & \textbf{99.8} \\
ToeSegmentation1 & 93.4 & 93.6 & 56.7 & \textbf{95.7}& 94.3 & 91.5 & 95.5 \\
ToeSegmentation2 & 92.8 & 93.5 & 66.6 & 94.9 & 93.3 & 92.3 & \textbf{96.0} \\
Trace & \textbf{100.0}& \textbf{100.0}& 95.4 & \textbf{100.0}& \textbf{100.0}& \textbf{100.0}& \textbf{100.0} \\
TwoLeadECG & 99.6 & 99.7 & 94.5 & 99.3 & 99.2 & 91.3 & \textbf{99.9} \\
TwoPatterns & \textbf{100.0}& \textbf{100.0}& 99.8 & 98.7 & 99.4 & 99.6 & \textbf{100.0} \\
UMD & 98.6 & 98.6 & \textbf{99.3}& 98.6 & 98.7 & 96.4 & 98.6 \\
UWaveGestureLibraryAll & 97.4 & \textbf{97.7}& 97.1 & 92.3 & 89.9 & 96.4 & 97.5 \\
UWaveGestureLibraryX & 87.0 & \textbf{87.5}& 81.6 & 83.8 & 82.1 & 83.9 & 86.0 \\
UWaveGestureLibraryY & 80.4 & \textbf{81.1}& 73.8 & 76.2 & 74.6 & 76.6 & 78.9 \\
UWaveGestureLibraryZ & 81.9 & \textbf{82.5}& 75.9 & 79.4 & 78.1 & 77.7 & 80.3 \\
Wafer & 99.9 & 99.9 & 99.6 & 99.8 & 99.7 & 99.7 & \textbf{100.0} \\
Wine & 93.5 & 94.1 & 94.2 & 94.4 & 93.7 & \textbf{95.0}& 92.5 \\
Worms & \textbf{77.4}& 77.1 & 59.6 & 71.1 & 73.8 & 67.4 & 75.8 \\
WormsTwoClass & 78.1 & 78.2 & 60.8 & 74.5 & 76.5 & 74.9 & \textbf{80.5} \\
Yoga & 91.0 & 91.7 & 89.0 & 86.8 & 88.1 & 88.7 & \textbf{93.0} \\
\end{longtable}
}

{\small
\setlength{\tabcolsep}{5pt}
\begin{longtable}{l ccc}
\caption{\textbf{UEA per-dataset results} (mean accuracy \%, 30 resamples). HC2 is blank for datasets outside the standard benchmark of~\citet{middlehurst2021hivecote2}. \textbf{Bold} indicates the best result per dataset (ties included). HC2 results were obtained from the published benchmark of~\citet{middlehurst2021hivecote2}; all other results were produced by the authors.}
\label{tab:uea_results}\\
\toprule
Dataset & $G{=}1$ & $G{=}10$ & HC2 \\
\midrule
\endfirsthead
\multicolumn{4}{c}{{\tablename\ \thetable{} -- continued}} \\
\toprule
Dataset & $G{=}1$ & $G{=}10$ & HC2 \\
\midrule
\endhead
\midrule \multicolumn{4}{r}{{Continued on next page}} \\
\endfoot
\bottomrule
\endlastfoot
AtrialFibrillation & 25.6 & 27.8 & \textbf{28.2} \\
BasicMotions & 98.3 & 98.6 & \textbf{98.9} \\
DuckDuckGeese & 43.7 & 48.8 & \textbf{49.9} \\
ERing & 96.9 & 97.4 & \textbf{98.5} \\
EigenWorms & 91.3 & \textbf{91.9}& 89.4 \\
Epilepsy & 99.3 & 99.6 & \textbf{99.8} \\
EthanolConcentration & 68.8 & 73.3 & \textbf{79.1} \\
FaceDetection & 64.2 & 68.9 & \textbf{71.3} \\
FingerMovements & 56.2 & \textbf{58.4}& 55.2 \\
HandMovementDirection & 38.7 & \textbf{46.4}& 39.8 \\
Heartbeat & 71.3 & 72.4 & \textbf{72.9} \\
InsectWingbeat & 28.7 & \textbf{38.8}&  \\
JapaneseVowels & 83.1 & \textbf{84.9}&  \\
MotorImagery & 51.4 & 51.7 & \textbf{53.2} \\
NATOPS & 88.2 & 88.9 & \textbf{89.2} \\
PEMS-SF & 92.9 & 96.6 & \textbf{99.8} \\
PenDigits & \textbf{99.7}& \textbf{99.7}& 99.6 \\
RacketSports & 91.3 & 92.6 & \textbf{93.0} \\
SelfRegulationSCP1 & 86.7 & \textbf{87.9}& \textbf{87.9} \\
SelfRegulationSCP2 & 51.6 & \textbf{53.4}& 50.5 \\
SpokenArabicDigits & 99.5 & \textbf{99.6}&  \\
StandWalkJump & \textbf{44.9}& 43.8 & 43.8 \\
UWaveGestureLibrary & 94.7 & \textbf{95.3}& 94.9 \\
\end{longtable}
}

\section{Excluded Datasets}
\label{app:excluded}

\tabpfn~v2.5 supports up to $C = 10$ classes and $n_\text{train} = 50{,}000$ training samples.
Of the 128~UCR datasets, 25 are excluded solely due to the class constraint ($C$ ranging from 11 to 60): Adiac ($C{=}37$), CricketX/Y/Z ($C{=}12$), Crop ($C{=}24$), EOGHorizontalSignal ($C{=}12$), EOGVerticalSignal ($C{=}12$), FaceAll ($C{=}14$), FacesUCR ($C{=}14$), FiftyWords ($C{=}50$), Fungi ($C{=}18$), GestureMidAirD1/D2/D3 ($C{=}26$), InsectWingbeatSound ($C{=}11$), NonInvasiveFetalECGThorax1/2 ($C{=}42$), Phoneme ($C{=}39$), PigAirwayPressure/ArtPressure/CVP ($C{=}52$), PLAID ($C{=}11$), ShapesAll ($C{=}60$), SwedishLeaf ($C{=}15$), and WordSynonyms ($C{=}25$).
Of the 30~UEA datasets, 7 are likewise excluded: ArticularyWordRecognition ($C{=}25$), CharacterTrajectories ($C{=}20$), Cricket ($C{=}12$), Handwriting ($C{=}26$), Libras ($C{=}15$), LSST ($C{=}14$), and PhonemeSpectra ($C{=}39$).
No dataset in either archive exceeds the training-size limit.

The class constraint could be overcome with standard multi-class decomposition strategies such as one-vs-one, one-vs-rest, or error-correcting output codes~\citep{dietterich1994ecoc}, each of which reduces a $C$-class problem to a collection of binary (or smaller-class) subproblems compatible with \tabpfn.
The training-size limit---not currently binding---could similarly be addressed through ensemble subsampling, where each of the $G$ \tabpfn instances operates on a random subsample of the training set, naturally extending the approach to larger datasets as tabular foundation models continue to scale.

\section{RocketPFN with TabPFN v3}
\label{app:tabpfn3}

\paragraph{Setup.}
\tabpfn~v3~\citep{grinsztajn2026tabpfn3technicalreport} was released after the experiments in the main text were completed.
We re-ran the two \tabpfn-dependent parts of our study, \tabpfn applied to flattened series and the full \ours pipeline, under \tabpfn~v3, keeping the feature extractor, hyperparameters, and 30-resample protocol of Section~\ref{sec:protocol} unchanged, and we compare on the same 92~UCR datasets used in the main text.
The larger class limit of \tabpfn~v3 additionally makes 26~datasets usable that \tabpfn~v2.5 cannot process ($C > 10$; see Appendix~\ref{app:excluded}), which we do not add here.

\paragraph{Temporal encoding is still necessary.}
\tabpfn~v3 applied to raw flattened series reaches a mean accuracy of only $0.844$, far below the encoded pipeline, and still scores at chance ($49.6$) on ShapeletSim despite v3's revised training prior.
The \rocket front-end remains essential.

\begin{figure}[htbp]
    \centering
    \includegraphics[width=0.82\linewidth]{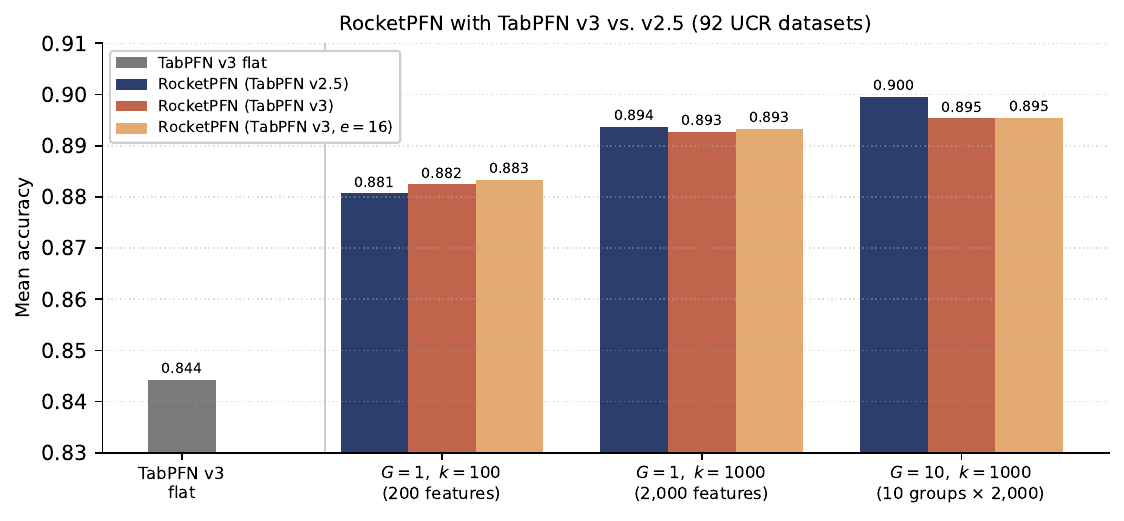}
    \caption{\textbf{\ours with \tabpfn~v3 vs.\ v2.5} (mean accuracy over 92~UCR datasets, 30-resample protocol).
    The leftmost bar is \tabpfn~v3 applied to flattened series, which stays far below.
    The remaining groups show \ours at three settings, comparing \tabpfn~v2.5 with \tabpfn~v3 under its automatic internal ensemble and under a fixed $e{=}16$.
    \tabpfn~v3 is ahead of v2.5 at the $200$-feature setting and behind it at the $2{,}000$-feature settings.}
    \label{fig:tabpfn3}
\end{figure}

\paragraph{\ours with \tabpfn~v3 vs.\ v2.5.}
Figure~\ref{fig:tabpfn3} compares the two \tabpfn versions inside \ours at three settings: a single group of $100$ kernels ($200$ features), a single group of $1{,}000$ kernels ($2{,}000$ features), and the default of ten groups of $1{,}000$ kernels each.
Each bar is the mean accuracy over the 92~UCR datasets.
The outcome splits by feature budget: \ours with \tabpfn~v3 beats v2.5 at the $200$-feature setting ($0.882$ vs.\ $0.881$), but loses at the $2{,}000$-feature settings, both with one group ($0.893$ vs.\ $0.894$) and with ten ($0.895$ vs.\ $0.900$).
Raising \tabpfn~v3's internal ensemble to $e{=}16$ does not reverse this: v3 stays ahead at $200$ features and behind at $2{,}000$.
We use \tabpfn~v2.5 throughout the main text.

\FloatBarrier

\end{document}